# MAXIMUM LIKELIHOOD AND MINIMUM ENTROPY IDENTIFICATION OF GRAMMARS[*]

P.Collet[1], A.Galves[2], A.Lopes[3]

*Abstract :  Using the Thermodynamic Formalism, we introduce a Gibbsian model for the identification of regular grammars based only on positive evidence. This model mimics the natural language acquisition procedure driven by prosody which is here represented by the thermodynamical potential. The statistical question we face is how to estimate the incidence matrix of a subshift of finite type from a sample produced by a Gibbs state whose potential is known. The model acquaints for both the robustness of the language acquisition procedure and language changes. The probabilistic approach we use avoids invoking ad-hoc restrictions as Berwick's Subset Principle.*

*key words :  grammars identification, Gibbs states, maximum likelihood, entropy.*
*AMS (MOS) Subject Classification : 58F11*
*Condensed title : Identification of grammars*

---

[*]  Work supported by FAPESP grant 90/3918-5 for Projeto Temático " Transição de Fase Dinâmica em Sistemas Evolutivos"
[1]  Centre de Physique Théorique, Ecole Polytechnique, 91128 Palaiseau Cedex (France). Laboratoire CNRS UPR 14, collet@orphee.polytechnique.fr.
[2]  Instituto de Matemática e Estatística, Universidade de São Paulo, BP 20570, 01452-970, São Paulo, SP, Brasil, galves@ime.usp.br
[3]  Instituto de Matemática, Universidade Federal do Rio Grande do Sul, 91500 Porto Alegre, RGS, Brasil, alopes@if1.ufrgs.br

# 1. INTRODUCTION

In this paper we present a simplified model of grammar identification, which tries to catch some of the main features of the process by which a child acquires a language. Formally this amounts to solve the following statistical question. In a subshift of finite type, how to infer the incidence matrix, given a finite sample chosen according to a Gibbs measure whose potential is known.

In this introduction we will sketch, in a very simplified way, how the problem is formulated from the point of view of Linguistics. In the next section we present the mathematical model and state the theorems. Readers who are only interested in the mathematical aspects of the problem should go directly to the next section.

In Chomsky's *Principles and Parameters* framework, the problem of understanding language acquisition can be roughly formulated in the following terms. A child has a genetic inherited *linguistic capacity* which makes him able to learn a language. This linguistic capacity is characterized by a finite set of constraints which distinguish natural languages among all the possible formal languages. This set of containts is what Chomsky calls the *Universal Grammar*. Any particular solution of these contraints is called a *grammar* and defines in a precise way a natural language. Therefore, "*learning a language*" is nothing but identifying an element in the set of natural grammars. We refer the reader to Chomsky 1986, for a comprehensive introduction to the Principles and Paramenters Model.

To identify the parental grammar a learning child is guided by the linguistic information available in his environment. Psycholinguists agree that corrections of wrong constructions do not play an important role in the learning process (cf. McNeill 1966). Therefore the model must use only positive evidence as a basis of inference.

The idea that the parental prosody helps the learning child to achieve his identification task appears recently in the linguistics litterature. Informally speaking, the prosody of a language is its characateristic *music*, which contains among other things, its typical stress and intonational patterns. Phonologists commonly accept the assumptiom that the prosody of a language depends on its syntax, even if a learning child acquires prosody before fixing his grammar. Therefore is natural to suppose that once acquired, prosody provides the learning child with hints about the parental grammar. This is the point of view we adopt here. We refer the reader to Galves and Galves 1993, where this point of view is applyed to a concrete linguistic situation.

An identification model must take into consideration the fact that languages change. Following Lightfoot 1979, grammatical changes occur during the acquisition process. From time to time a generation of learning children chose a grammar which is different from the parental one. It is has been argued that some of those changes may have been induced by a former prosodic change. Therefore the model must account both for the robustness of the acquisition process, and for the possibility of misidentification driven by some particular prosodic choices.

The learning process can be naturally considered as a random process. The sequence of sentences the learning child receives from his parents does not follow any detemininistic order (parents do not follow any kind of "*manual*" to teach a language to their child). This random process is stationary in time and its law depends on the the parental syntax and prosody. Therefore the basis of a reasonable model of language acquisition must be a probability measure having the language as its *sample space* and having the syntax and the prosody as parameters. The Thermodynamical Formalism provides a natural way to express this. We refer the reader to Ruelle 1978 for a general presentation of the Thermodynamical Formalism.

The issue we consider here was firt addressed in a rigorous mathematical way in Gold 1967 through the *identification in the limit model*. This was not a probabilistic model and did not take into consideration prosody as an element playing a role in the identification process. This model ended by a *constat d'échec*. The identification in the limit procedure never converges to a unique grammar. To overcome this failure, it has been suggested in the linguistic litterature (Berwick 1985, following Angluin 1980) that an extra principle should be taken into consideration, the so called *Subset Principle*. However a probabilistic point of view like the one we adopt here solves the problem in a more natural way.

In the present paper we restrict our study to what in Chomsky's hierarchy is called *regular grammars* (cf. Chomsky 1963). Since Chomsky 1956, it is well known that regular grammars are just too a rough concept to catch the subtle properties of natural languages. However we do believe that our mathematical results express in a simplified way part of the real story.

## 2. DEFINITIONS AND RESULTS

A lexicon is a finite set $\Lambda$. A grammar $G$ acting on the lexicon $\Lambda$ is a matrix indexed by $\Lambda$ and with entries equal to 0 or 1. We will only consider irreducible and aperiodic matrices, i.e. there is an integer $k$ such that all the entries of the matrix $G^k$ are nonzero. These matrices are also called primitive in the literature (see Horn and Johnson 1985). We will denote by $\mathcal{G}$ the set of all such grammars.

The language generated by $G$ is the set

$$L(G) = \{(x_0, \cdots), \ x_j \in \Lambda, \ G_{x_j, x_{j+1}} = 1, \ j \geq 0\} \ .$$

We introduce a partial order in G in the following way. If $G$ and $G'$ belong to $\mathcal{G}$ we say that $G < G'$ if for all pairs $(x, y) \in \Lambda^2$ we have $G(x, y) \leq G'(x, y)$ and the inequality is strict for at least one pair.

Note that $G < G'$ is equivalent to $L(G) \subset \neq L(G')$.

Let the sampler $S_n$ be the map from $\Lambda^{\mathbb{N}}$ to $\Lambda^n$ which gives the first $n$ symbols of an infinite string.

We are interested in the problem of identifying a grammar in $\mathcal{G}$ given a sample produced by $S_n$ acting on the language defined by a fixed but unknown grammar. It is natural to consider $S_n$ as a random variable. In order to make this precise, for every $G \in \mathcal{G}$ we introduce a probability measure on $L(G)$ equiped with the usual $\sigma$-field induced by the product $\sigma$-algebra.

Since the grammar $G$ is the unknown in our problem, we need a canonical construction of the probability measure. A natural way of doing this is to fix a real valued Hölder continuous function $\phi$ on $\Lambda^{\mathbb{N}}$, and to associate to any grammar $G \in \mathcal{G}$ the Gibbs state with potential $\phi$. We will denote this Gibbs measure by $\mu_\phi^G$. The classical references to Gibbs measures are Bowen 1975 and Ruelle 1978. An extensive and up-to date reference is Parry and Pollicott 1990. In particular the reader will find there a proof of the existence and unicity of the measure $\mu_\phi^G$ for $\phi$ in $C^\alpha$, the Banach space of $\alpha$-Hölder continuous functions, equiped with the usual $C^\alpha$ norm, for any fixed $\alpha$.

We recall that $\mu_\phi^G$ is the unique measure such that there is a positive constant $C > 1$ such that for any element $\underline{x}$ of $L(G)$ and for any integer $n$ we have

$$C^{-1} \leq \frac{\mu_\phi^G(\{\underline{y} \,:\, S_n(\underline{y}) = S_n(\underline{x})\})}{e^{-nP + \sum_{j=0}^{n-1} \phi(\sigma^j \underline{x})}} \leq C \,, \tag{1}$$

where $P = P(\phi, G)$ is the pressure associated to the potential $\phi$ on $L(G)$ (see Theorem 1.2 in Bowen 1975).

¿From now on we will use the shorthand notation $[S_n(\underline{x})]$ to denote the cylindrical set $\{\underline{y} \,:\, S_n(\underline{y}) = (x_0, \cdots, x_{n-1})\}$.

For a fixed $\phi$, and for any string $\underline{x}$, we define the sequence of Maximum Likelihood subsets $\mathcal{M}_\phi^n(\underline{x})$ ($n \geq 1$) of $\mathcal{G}$ by

$$\mathcal{M}_\phi^n(\underline{x}) = \left\{ G \,:\, \mu_\phi^G([S_n(\underline{x})]) = \max_{G' \in \mathcal{G}} \mu_\phi^{G'}([S_n(\underline{x})]) \right\} \,.$$

We can now define the **Maximum Likelihood Identification Procedure**: for any $n$, given $\phi$ and the sample $S_n(\underline{x})$ the learner chooses a grammar belonging to $\mathcal{M}_\phi^n(\underline{x})$. This procedure is non ambiguous if $\mathcal{M}_\phi^n(\underline{x})$ is a singleton.

Our first identification Theorem says that the Maximum Likelihood Procedure always identifies the departure grammar in the limit as $n$ diverges.

**Theorem A.** *For any potential $\phi$ and any grammar $G$ the Maximal Likelihood sets $\mathcal{M}_\phi^n(\underline{x})$ converges to $\{G\}$ for $\mu_\phi^G$ almost all choices of $\underline{x}$, as $n$ diverges.*

The above Theorem accounts for the robustness of the learning process. A child which uses the Maximum Likelihood Procedures to identify the parental grammar succeeds using a finite sample of positive evidences.

Nevertheless, languages change. Since in a natural language acquisition situation the identification is done with a fixed $n$ which is biologically defined, it seems reasonable to think of a model in which the Identification Procedure is based on a large but finite sample. In this case, the Maximum Likelihood Procedure can given an unambiguous answer which is neverthless different from the departure grammar. This is summarized in the following Proposition, which is trivial and will not be proved.

**Proposition B.** *For any $n \geq 1$, $G$ and $G'$ in $\mathcal{G}$, such that $G > G'$, and any $\epsilon \geq 0$ there exists a Hölder continuous potential $\phi$ such that*

$$\mu_\phi^G(\{\underline{x} \,:\, G' \in \mathcal{M}_\phi^n(\underline{x}) \text{ but } G \notin \mathcal{M}_\phi^n(\underline{x})\}) \geq 1 - \epsilon \,.$$

This model is not satisfactory, since it only describes changes leading to smaller grammars (*i.e.* grammars which allow less transitions than the parental one).

Trying to improve the model, we introduce a new procedure which coincides with the Maximum Likelihood aproach in most cases but nevertheless even in the limit of diverging $n$ can lead to a new grammar, which can be strictly greater than the original one. The next two Theorems show how this may occur under a Minimum Entropy Identification Procedure.

Given the Gibbs state $\mu_\phi^G$, let $h(\mu_\phi^G)$ denote its Kolmogorov-Sinai entropy (see Bowen 1975 for the definition).

The Shannon-McMillan-Breiman Theorem says that the $\mu_\phi^G$ measure of a cylindrical set $[S_n(\underline{x})]$ is tipically of order $e^{-nh(\mu_\phi^G)}$. This suggests that a Minimum Entropy Criterium could be used instead of the Maximum Likelihood Procedure we have just described. As a matter of fact Theorem C bellow shows that both approachs coincide for potentials which are close to the null potential, *i.e.* potentials belonging to

$$\mathcal{O}_r = \{\phi \ : \ \|\phi\|_{C^\alpha} < r\}$$

the $C^\alpha$ ball with radius $r$ centered at the null potential, where $r$ is sufficiently small.

We define the Minimum Entropy Subset $\mathcal{E}_\phi^n(\underline{x})$ by

$$\mathcal{E}_\phi^n(\underline{x}) = \left\{G \ : \ [S_n(\underline{x})] \subset L(G) \quad \text{and} \quad h(\mu_\phi^G) \ \text{is minimal}\right\}.$$

We may now introduce the **Minimum Entropy Identification Procedure**. Given $\phi$, $\underline{x}$, and $n$ the learner chooses a grammar belonging to $\mathcal{E}_\phi^n(\underline{x})$.

**Theorem C.** *There exists a positive real number $r$ such that for any potential $\phi$ in $\mathcal{O}_r$ and any grammar $G$ the Minimum Entropy sets $\mathcal{E}_\phi^n(\underline{x})$ converge to $\{G\}$ for $\mu_\phi^G$ almost all choices of $\underline{x}$, as $n$ diverges.*

**Theorem D.** *For any $G$ and $G'$ in $\mathcal{G}$, such that $G < G'$, there exists a Hölder continuous potential $\phi$ such that for $\mu_\phi^G$ almost every $\underline{x}$ and for any $n$ large enough $G' \in \mathcal{E}_\phi^n(\underline{x})$ but $G \notin \mathcal{E}_\phi^n(\underline{x})$.*

## 3. PROOF OF THEOREM A.

Theorem A will be proved as soon as we show that for any potential $\phi$ and $n$ large enough the Maximum Likelihood set excludes both grammars which have an entry smaller than the original grammar, and grammars which are strictly larger. This is done in the next two lemmata.

**Lemma 1.** *Let $G$ and $G'$ be two grammars such that $G' < G$. Then*

$$\lim_{n \to \infty} \mu_\phi^G(\{\underline{x} : G' \in \mathcal{M}_\phi^n(\underline{x})\}) = 0 \ .$$

**Proof.** This follows directly from the Ergodic Theorem, which says that every event which has positive probability, does indeed occur.

**Lemma 2.** *For any $\phi \in C^\alpha$ there is an integer $n(\phi)$ such that for any pair $G$ and $G'$ of grammars such that $G < G'$, then for any string $\underline{x} \in L(G)$ we have*

$$\mu_\phi^{G'}([S_n(\underline{x})]) < \mu_\phi^G([S_n(\underline{x})])$$

*for all $n \geq n(\phi)$.*

**Proof.** From inequality (1) and the finiteness of $\mathcal{G}$, it follows that there is a constant $C$ (which depends only on $\phi$) such that for any integer $n$ and any string $\underline{x} \in L(G)$ we have

$$\mu_\phi^G([S_n(\underline{x})]) > C e^{n(P(\phi,G) - P(\phi,G'))} \mu_\phi^{G'}([S_n(\underline{x})]) \ .$$

To conclude the proof it is enough to use the following proposition.

**Proposition 3.** *Let $\phi \in C^\alpha(\Lambda^{\mathbb{N}})$, then if $G < G'$, we have*

$$P(\phi, G) < P(\phi, G') \ .$$

Note in particular that in the above proposition the inequality is strict.

**Proof.** We recall (see Bowen 1975) that the pressure of $\phi$ is the logarithm of the largest eigenvalue of the transfer operator defined on $C^\beta(L(G))$ ($0 < \beta < \alpha < 1$) by

$$\mathcal{L}_G^\phi \psi(x_0, x_1, \cdots) = \sum_{x \in \Lambda, \ G(x, x_0) = 1} e^{\phi(x, x_0, x_1, \cdots)} \psi(x, x_0, x_1, \cdots) \ .$$

In what follows, if there is no danger of confusion, we shall use the shorter notation $\mathcal{L}_G$ instead of $\mathcal{L}_G^\phi$.

Note that if $G < G'$ and $\psi$ is non-negative then $\mathcal{L}_G \psi \leq \mathcal{L}_{G'} \psi$.

We recall also that in the Banach space $C^\beta(L(G))$, the operator $\mathcal{L}_G$ has a simple isolated eigenvalue denoted below by $\lambda(G)$ which is real and positive, the rest of the spectrum is contained in a disk centered at the origin with radius strictly smaller than this number $\lambda(G)$. It is known that $\log(\lambda(G)) = P(\phi, G)$.

Moreover, the associated eigenvector $\psi_G$ is positive and bounded below away from zero, and the associated eigencovector $\alpha_G$ is a positive measure. Similar results hold of course for $G'$, and we are going to prove that $\lambda(G) < \lambda(G')$.

This will follow from the eigenvalue equation for $G'$ considered on $L(G)$. We have indeed for $\underline{x} \in L(G)$ the relation

$$\lambda(G')\psi_{G'}(\underline{x}) = \mathcal{L}_{G'}\psi_{G'}(\underline{x}) = \mathcal{L}_G \psi_{G'}(\underline{x}) + r_{G'}(\underline{x}) ,$$

where

$$r_{G'}(\underline{x}) = \sum_{x \in \Lambda, G(x,x_0)=0, G'(x,x_0)=1} e^{\phi(x,x_0,x_1,\cdots)} \psi_{G'}(x, x_0, x_1, \cdots) .$$

Note however that for some choice of $x_0$, $r_{G'}(\underline{x})$ may be equal to zero. In order to deal with this problem we iterate the above equality $n$ times where $n$ is the smallest integer such that all entries of the matrix $G^n$ are non zero (this number is finite since $G$ is irreducible and aperiodic). We obtain since $r_{G'} \geq 0$ and $\mathcal{L}_G$ is positivity preserving

$$\lambda(G')^n \psi_{G'}(\underline{x}) \geq \mathcal{L}_G^n \psi_{G'}(\underline{x}) + \mathcal{L}_G^{n-1} r_{G'}(\underline{x}) , \qquad (2)$$

where

$$\mathcal{L}_G^{n-1} r_{G'}(\underline{x}) = \sum_{\substack{x_{-n-1}, \cdots x_{-1} \in \Lambda \\ G'(x_{-n-1}, x_{-n}) = G(x_{-n}, x_{-n+1}) = \cdots = G(x_{-1}, x_0) = 1}} \prod_{j=0}^{n} e^{\phi(x_{-n-1+j}, x_{-n+j}, \cdots)} \psi_{G'}(x_{-n-1}, x_{-n}, \cdots) .$$

Form our choice of $n$, and the aperiodicity of $G$, we conclude that for any $\underline{x} \in L(G)$, there is at least one term in the above sum which is non zero. Moreover, since the function $\psi_{G'}$ is bounded below away from zero (and bounded above since it is continuous), we derive that there is a number $\eta > 0$ such that

$$\mathcal{L}_G^{n-1} r_{G'}(\underline{x}) \geq \eta \psi_{G'}(\underline{x}) ,$$

which implies by (2) that

$$(\lambda(G')^n - \eta)\psi_{G'} \geq \mathcal{L}_G^n \psi_{G'}$$

on $L(G)$. Since the eigencovector $\alpha_G$ is a positive measure, and since $\psi_{G'}$ is stricly positive, we have $\alpha_G(\psi_{G'}) > 0$. Therefore if we apply $\alpha_G$ to the two members of the above inequality, we get

$$\lambda(G')^n - \eta \geq \lambda(G)^n .$$

This implies $P(\phi, G') = \log \lambda(G') > \log \lambda(G) = P(\phi, G)$.

## 4. PROOF OF THEOREM C

Let $\mu_0^G$ be the Gibbs state associated to the null potential. We will first prove that the Minimum Entropy set converges to the original grammar with probability 1 with respect to $\mu_0^G$. Theorem C will then follow by continuity of the pressure as a function of the potential (cf Parry and Pollicott 1990), Theorem A and Lemma 7.

We recall that for a matrix $G \in \mathcal{G}$, the eigenvalue with largest modulus is simple and positive. The associated eigenvalue is the exponential of the topological entropy $h_{\text{top}}(G)$ of the Markov shift, and the associated eigenvalue is a vector with strictly positive entries (see Horn and Johnson 1985). In the notation of the previous section this corresponds to the null potential $\phi = 0$ (see Bowen 1975).

We now observe that if $G < G'$, there is a matrix $R$ with entries equal to 0 or 1 such that $G' = G + R$. However, if $h_{\text{top}}(G) = h_{\text{top}}(G')$, it follows form exercise 8.4.15 in Horn and Johnson 1985 that $R = 0$, i.e. $G' = G$.

The proof of Theorem C is a direct consequence of the following lemmata.

**Lemma 4.** *For any $G \in \mathcal{G}$ and for $\mu_0^G$ almost any $\underline{x} \in L(G)$*

$$\lim_{n \to \infty} \frac{\log \mu_0^G([S_n(\underline{x})])}{n} = -h_{top}(G)$$

**Proof.** The result follows directly from the Shannon-McMillan-Breiman Theorem.

**Lemma 5.** *Let $G$ and $G'$ be two elements of $\mathcal{G}$ with the same topological entropy. If $L(G) \subset L(G')$, then $G = G'$.*

**Proof.** Let $\mathcal{C}$ be the cone of vectors in $\mathbb{R}^\theta$ with positive coordinates, where $\theta$ denotes the cardinal of the lexicon $\Lambda$. We have that for any integer $n$

$$G^n \mathcal{C} \subset \mathcal{C} \quad \text{and} \quad {G'}^n \mathcal{C} \subset \mathcal{C} .$$

We will denote by $p$ (or $p^G$) the eigenvector of $G$ corresponding to the largest positive eigenvalue $\lambda = e^{h_{top}(G)}$, normalized by the condition

$$\sum_{i \in \Lambda} p_i^G = 1 .$$

We also denote by $\lambda'$ the corresponding eigenvalue of $G'$, and our hypothesis implies that $\lambda = \lambda'$. We will derive a contradiction from the assertion that

$$L(G) \subset \neq L(G') .$$

Since however $L(G) \subset L(G')$, we have $G' = G + B$ where $B$ is a matrix of dimension $\theta$ with entries equal to zero or one and with at least one non zero entry. We observe that due to the irreducible and aperiodic property, there is an integer $m$ such that for any pair of indices $(i, j)$

$$G_{i,j}^m > 0 .$$

We now fix $n = m + 1$, and conclude that for the order among vectors of $\mathbb{R}^\theta$ associated to the cone $\mathcal{C}$

$$G'^{m+1} p > G^{m+1} p + G^m B p \ .$$

We recall that $v_1 > v_2$ iff $v_1 - v_2 \in \mathcal{C}$ (see Bowen 1975). ¿From our above choice of $m$, since $B$ has at least one non zero entry and since the components of $p$ are strictly positive, we conclude that there is a strictly positive number $\eta$ such that

$$G'^{m+1} p > (\lambda^{m+1} + \eta) p \ .$$

Since $G'$ maps the cone $\mathcal{C}$ into itself we have for any integer $k$

$$G'^{k(m+1)} p > (\lambda^{m+1} + \eta)^k p \ ,$$

which implies $\lambda' > \lambda$, a contradiction.

**Lemma 6.** *There is a positive real number $r$ such that if $\phi \in \mathcal{O}_r$, the map $G \to h(\mu_\phi^G)$ is strictly monotone increasing.*

**Proof.** The result follows from the continuity of $h(\mu_\phi^G)$ with respect to $\phi$ for $G \in \mathcal{G}$ (cf Parry and Pollicott 1990) and Lemma 5.

**Lemma 7.** *There is a positive real number $r$ such that for any $\phi \in \mathcal{O}_r$ and for any $G \in \mathcal{G}$,*

$$\lim_{n \to \infty} \mu_\phi^G(\{\underline{x} \ : \ \mathcal{M}_\phi^n(\underline{x}) = \mathcal{E}_\phi^n(\underline{x}) = \{G\}\}) = 1 \ .$$

**Proof.** The result follows from Lemma 6.

Theorem C now follows from Theorem A and Lemma 7.

# 5. PROOF OF THEOREM D

The proof of Theorem D follows from Lemma 8.

**Lemma 8.** *Given two grammars $G$ and $G'$ in $\mathcal{G}$ with $G < G'$, there exists a Hölder continuous potential $\phi$ such that*

$$h(\mu_\phi^G) > h(\mu_\phi^{G'}) \ .$$

**Proof.** The idea of the proof is to find a potential such that the Gibbs state will look like the invariant measure supported by a periodic orbit (which has of course zero entropy). As in the previous theorem, the matrix $G'$ has at least one more entry equal to 1 than the matrix $G$. Therefore we can find a periodic orbit for the subshift of $G'$ which is not admissible for $G$. Let $(y_0, \cdots, y_{q-1})$ be such a periodic orbit with minimal period $q$. In other words, we have

$$G'_{y_j, y_{j+1}} = 1 \quad \text{for} \quad 0 \le j \le q-1 \ , \qquad G'_{y_q, y_1} = 1$$

and we can also assume that

$$G_{y_1, y_2} = 0 \ .$$

We now define the function $\phi$ as follows. Let $E$ be a positive number to be fixed later on, we set $\phi(y_0, \cdots, y_q) = E$ if $y_q = y_0$, and there is an integer $0 \le l \le q-1$ such that for any integer $0 \le j \le q-1$, $y_j = y_{j+l \pmod q}$, and $\phi(y_0, \cdots, y_q) = 0$ otherwise. Note that since $\phi$ is a cylindrical function, it is Hölder continuous.

On the set $L(G)$ which is the phase space of the subshift associated to $G$, the function $\phi$ is equal to 0. In this case, the corresponding Gibbs measure has maximal Kolmogorov-Sinai entropy, i.e. $h(\mu_\phi^G) = h_{\text{top}}(G)$.

We now consider the Gibbs state on $L(G')$. We observe that for a fixed positive number $\beta$, the transfer operator $\mathcal{L}_{G'}^{\beta\phi}$ associated to $\beta\phi$ on the subshift of $G'$ and given by

$$\mathcal{L}_{G'}^{\beta\phi}\psi(x_1, \cdots) = \sum_{x_0, G'_{x_0, x_1}=1} e^{\beta\phi(x_0, \cdots, x_q)} \psi(x_0, \cdots) \ ,$$

maps the space of cylindrical functions of the first $q$ variables into itself (we will of course only consider admissible sets of $q$ variables with respect to the matrix $G'$). This is a finite dimensional subspace of the space of Hölder continuous function, and if we find in this subspace a positive eigenvalue with a strictly positive eigenvector, by uniqueness this eigenvalue must be the exponential of the pressure of $\beta\phi$.

From now on it will be more convenient to use a matrix notation. Let $m$ denote the number of sequences of length $q$ admissible by $G'$. The real valued cylindrical functions which depend only on the first $q$ symbols form a real vector space of dimension $m$. It is easy to verify that in this space the transfer operator can be represented by a matrix $M_{\beta\phi}$ which takes the following form

$$M_{\beta\phi} = z^{-1}(M_0 + zM_1)$$

where $z = \exp(-\beta E)$, $M_0$ and $M_1$ are matrices with entries equal to 0 or 1. Moreover, there is a basis of the space $\mathbb{R}^m$ denoted by $e_0, \cdots, e_{m-1}$ such that

$$M_0 e_j = e_{j+1} \pmod{q} \quad \text{for} \quad 0 \leq j \leq q-1$$
$$M_0 e_j = 0 \quad \text{else}.$$

Note that $M_0$ is not primitive, but its spectrum is composed of the eigenvalue 0 with multiplicity $m - q$ and the $q^{\text{th}}$ root of unity which are simple eigenvalues. In particular, 1 is a simple eigenvalue. By analytic perturbation theory (see Parry and Pollicott 1990), we conclude that for $z$ small enough, the matrix $zM_{\beta\phi}$ has a simple eigenvalue $\lambda(z)$ which is an analytic function of $z$ which tends to 1 if $z \to 0$. We also know that the matrix $zM_{\beta\phi}$ is such that all it's entries are non negative, and moreover there is a power of this matrix with all it's entries strictly positive. This matrix has therefore a real positive eigenvalue which is simple and is also the unique point in the spectrum with maximum modulus. The associated eigenvector has strictly positive coordinates. For $z$ small enough we conclude that this point must be $\lambda(z)$. If we denote by $P(z)$ the function $\log \lambda(z)$ we have

$$h(\mu_\phi^{G'}) = \left(P + Ez\frac{dP}{dz}\right)_{z=e^{-E}} = \log \lambda(e^{-E}) + Ee^{-E}\frac{\lambda'(e^{-E})}{\lambda(e^{-E})},$$

and this number tends to zero if $E$ diverges. Therefore, for $E$ large enough, it will be smaller than $h_{\text{top}}(G)$ and the theorem is proven.

**Acknowledgements**. We thank Marzio Cassandro for helpful discussions. We are particularly grateful to Charlotte Galves for patiently introducing us to the problem and for pointing out the important relationship between prosody and syntax.